\documentclass[conference]{IEEEtran}
\providecommand{\IEEEoverridecommandlockouts}{}
\IEEEoverridecommandlockouts
\usepackage{cite}
\usepackage{amsmath,amssymb,amsfonts}
\usepackage{graphicx}
\usepackage{float}
\usepackage{stfloats}
\usepackage{booktabs}
\usepackage{xcolor}
\usepackage{url}
\usepackage{pifont}
\usepackage[hidelinks]{hyperref}

\DeclareMathOperator*{\argmin}{arg\,min}

\newcommand{\fieldname}[1]{\texttt{#1}}
\newcommand{\cmark}{{\color{green!55!black}\ding{51}}}
\newcommand{\xmark}{{\color{red!70!black}\ding{55}}}
\newcommand{\pmark}{{\color{orange!80!black}$\sim$}}

\begin{document}

\title{Vision--Language--Motion Maps: An Open-Vocabulary, \\ Uncertainty-Aware, Queryable Motion Attribute for 3D Scene Maps}

\author{Dibyendu Ghosh and Ayushi Shakya\\ Rekise Marine, Bangalore, India\\ \texttt{dibyendu35@gmail.com}}

\maketitle
\flushbottom

\begin{abstract}
Open-vocabulary 3D maps let robots answer language queries about \emph{what} and \emph{where},
but they assume a static world and cannot answer queries about how scene elements \emph{behave}.
We introduce Vision--Language--Motion Maps (VLMM), an open-vocabulary, language-queryable
3D map---queried through a rule-based intent router over open-vocabulary object nouns, not a
general natural-language interface---in which each element carries a \emph{fused motion attribute}: a VLM/LLM semantic
\emph{movability prior} combined with geometrically \emph{observed} cross-frame motion, together
with a \emph{per-element uncertainty}. Queries reduce to attribute filters that distinguish what
\emph{has been seen to move}, what \emph{could move but has not}, and what \emph{stays still}.
On a controlled simulator benchmark with exact ground truth (AI2-THOR, three scene types) we show
through ablation that the schema fields are non-substitutable: a semantic-only baseline fails motion
queries even with strong features, and neither motion field substitutes for the other (the prior
cannot answer ``what is moving,'' observed motion cannot answer ``what could move''). On real dynamic RGB-D (TUM and Bonn, six
sequences) we show the uncertainty channel---our key difference from prior fused-motion work---
consistently improves moving-vs-static average precision and reduces false motion flags, and that
it is robust to estimated (noisy) poses. The raw confidence is not calibrated, but post-hoc isotonic
calibration reaches an expected calibration error of $0.10$. VLMM is a representation contribution:
the closest prior maps each lack at least one of the four properties---open-vocabulary,
language-queryable, fused prior-and-observed motion, and per-element uncertainty---that our
combination provides.
\end{abstract}

\vspace{2pt}
\vspace{2pt}

\section{Introduction}
A mobile robot that is told ``the door that opens,'' ``things I could move,'' or ``where it stays
still'' needs more than a static semantic map: it must know how scene elements \emph{behave}---a
property existing maps largely do not represent, as we now survey.

\smallskip\noindent\textbf{Static open-vocabulary maps.} VLMaps~\cite{huang2023vlmaps},
ConceptFusion~\cite{jatavallabhula2023conceptfusion}, ConceptGraphs~\cite{gu2024conceptgraphs}, and
HOV-SG~\cite{werby2024hovsg} attach open-vocabulary visual--language features to top-down grids,
voxels, or object nodes and answer natural-language spatial queries, but assume a static world and
store no motion; VLMaps itself lists extending to ``dynamic objects and moving humans'' as future work.

\noindent\textbf{``Dynamic'' open-vocabulary maps as map maintenance.} DynaMem~\cite{dynamem2024} and
DovSG~\cite{dovsg2025} handle change by add/remove and re-perception, without a persistent,
uncertainty-bearing, queryable motion field. DualMap~\cite{dualmap2025}---closest to us---carries a
CLIP anchor-vs-volatile movability \emph{prior} and re-detects objects, but does \emph{not} fuse the
prior with measured geometric motion and stores no motion uncertainty.

\noindent\textbf{Geometric and fused motion.} Khronos~\cite{schmid2024khronos} builds a 4D
metric-semantic map from geometric motion detection but has no semantic prior, no stored per-element
uncertainty, and no natural-language motion query. Dewan~\textit{et al.}~\cite{dewan2017deep} fuse a
learned objectness prior with observed LiDAR motion in a Bayes filter (non-movable/movable/dynamic
classes with a per-point belief), but are closed-set and not language-queryable.
Voxeland~\cite{voxeland2024} models per-instance semantic uncertainty but is static.
Table~\ref{tab:capability} summarizes: \emph{no prior system provides all four} of
\{open-vocabulary, language-queryable, fused prior$\times$observed motion, per-element uncertainty\}.
We credit Dewan's Bayesian belief and DualMap's prior explicitly, so the distinction rests on the
\emph{combination}, not a single column.

\smallskip\noindent We propose \textbf{Vision--Language--Motion Maps (VLMM)}, whose map elements carry the schema
\begin{equation*}\small
\begin{aligned}
\langle\, &\text{position},\ \fieldname{semantic\_feat},\ \fieldname{observed\_motion}(\!+\!\text{conf}),\\[-1pt]
          &\fieldname{movability\_prior},\ \fieldname{motion\_class}(\!+\!\text{conf}) \,\rangle .
\end{aligned}
\end{equation*}
Our contributions are: (i) a \emph{queryable, uncertainty-aware fused motion attribute} for
open-vocabulary 3D maps; (ii) a geometric observed-motion channel that propagates range-dependent
depth covariance into a Mahalanobis motion score, and that is robust to estimated poses via
ego-motion refinement; (iii) a language-query interface that routes each query to a single schema
field; and (iv) an evaluation that shows the schema fields are non-substitutable (no single field,
including a strong semantic feature, answers the motion queries) across genuinely different
scenes with exact ground truth, plus a characterization of where the contribution is strong
(relative ablation, multi-scene uncertainty) and where it is not (absolute real-world accuracy,
calibration).

\begin{table}[t]
\centering
\caption{\footnotesize Capability comparison (\cmark\,yes, \pmark\,partial, \xmark\,no).}
\label{tab:capability}
\footnotesize
\setlength{\tabcolsep}{3pt}
\resizebox{0.98\columnwidth}{!}{
\begin{tabular}{lccccc}
\toprule
System & open- & NL- & queryable & fused & per-elem.\\
 & vocab & query & motion & prior$\times$obs & uncert.\\
\midrule
VLMaps~\cite{huang2023vlmaps}        & \cmark & \cmark & \xmark & \xmark & \xmark\\
ConceptGraphs~\cite{gu2024conceptgraphs} & \cmark & \cmark & \xmark & \xmark & \xmark\\
Khronos~\cite{schmid2024khronos}     & \pmark$^1$ & \xmark & \xmark & \xmark & \xmark\\
Dewan'17~\cite{dewan2017deep}        & \xmark & \xmark & \xmark & \cmark & \pmark$^2$\\
DualMap~\cite{dualmap2025}           & \cmark & \cmark & \pmark$^3$ & \xmark & \xmark\\
\textbf{VLMM (ours)}                 & \cmark & \pmark$^4$ & \cmark & \cmark & \cmark\\
\bottomrule
\end{tabular}}
\\[2pt]
{\scriptsize $^1$open-set SAM+CLIP variant; no NL motion query.\\
$^2$per-point Bayes belief, closed-set.\quad
$^3$anchor/volatile state, not a queryable motion attribute.\\
$^4$motion attribute is language-queryable, but via a rule-based intent router
($80\%$ of paraphrases, fails negation); LLM parser is future work.}
\end{table}

\section{Method}
\begin{figure*}[t]
\centering
\includegraphics[width=0.97\textwidth]{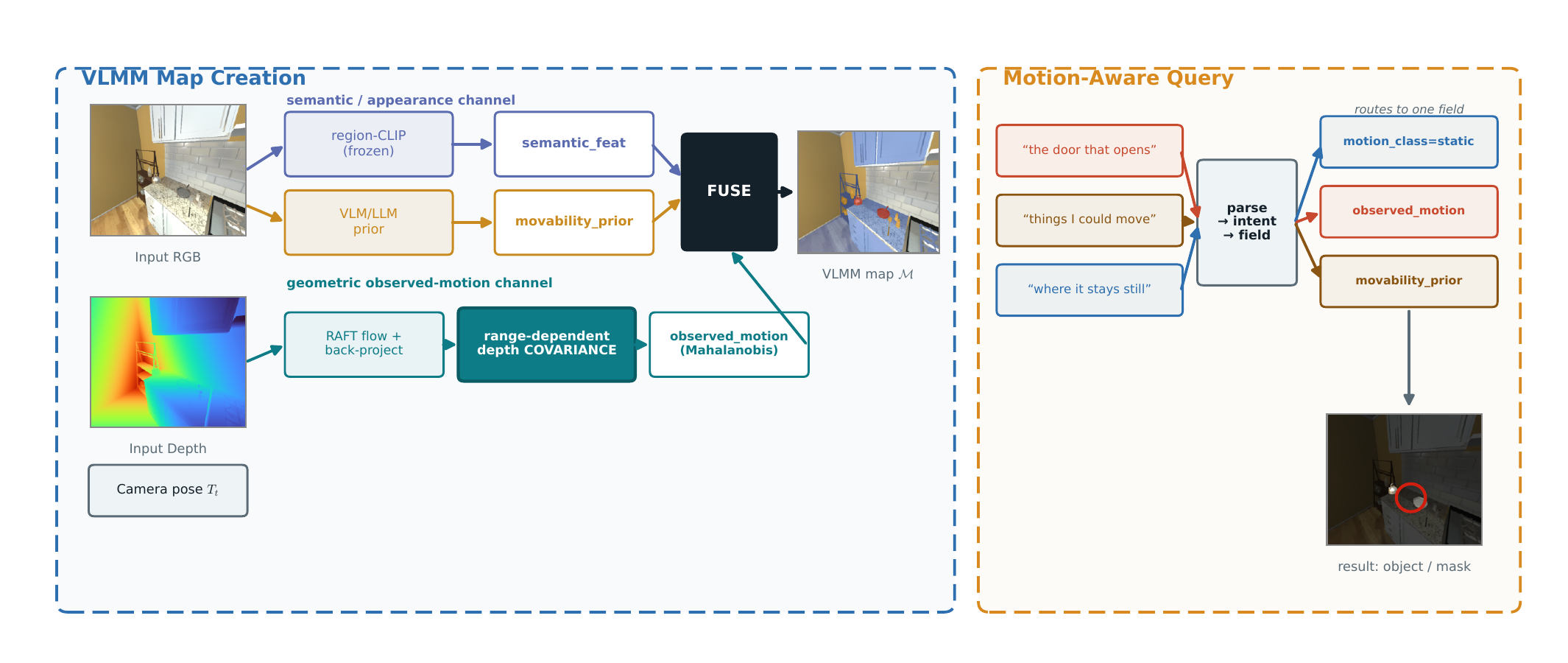}
\vspace{-10pt}
\caption{VLMM builds each element from two channels and routes each query to one field. The
range-dependent depth-covariance box (teal) is what neither prior system represents in an
open-vocabulary, language-queryable map; it makes the motion attribute uncertainty-aware.}
\label{fig:arch}
\end{figure*}

VLMM represents a scene as a set of object-instance elements, each carrying a fused,
uncertainty-aware motion attribute, and answers a natural-language query by routing it to a single
schema field. Each element is built from two complementary channels: a semantic/appearance channel
that yields an open-vocabulary feature and a movability prior, and a geometric channel that measures
observed motion together with a propagated uncertainty. A confidence-aware rule fuses the two into a
single \fieldname{motion\_class}, after which queries reduce to attribute filters over the schema
(Fig.~\ref{fig:arch}). We first fix notation, then describe the two channels, the fusion rule, and
the query interface.

\begin{figure*}[tbp]
\centering
\includegraphics[width=0.97\textwidth]{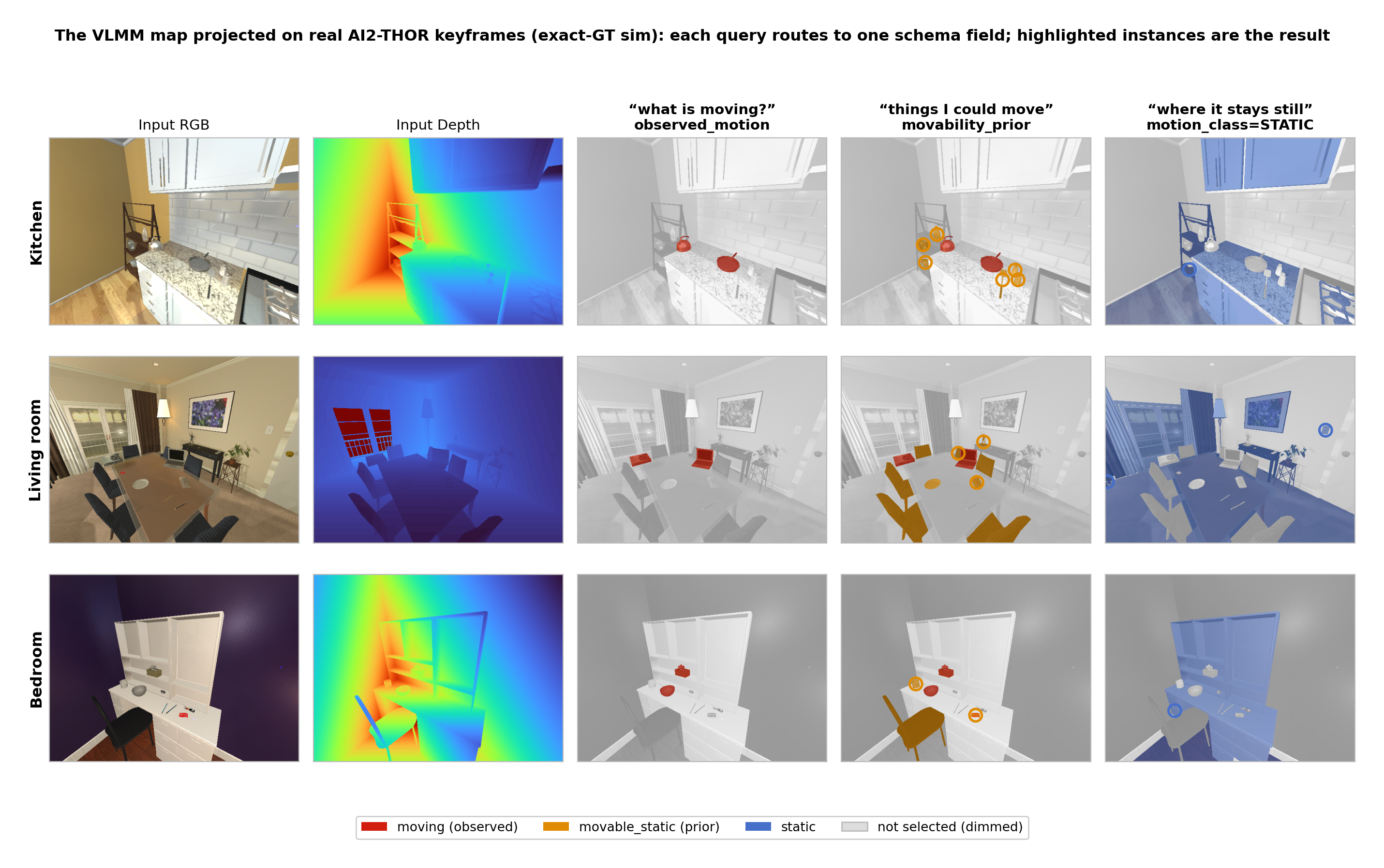}
\vspace{-10pt}
\caption{\footnotesize The VLMM map projected on real AI2-THOR keyframes (exact-GT sim);
rows are three scenes. The first two columns are the input RGB and depth; the next three are
natural-language queries, each routed to a single schema field: ``what is moving''
($\to\!$\fieldname{observed\_motion}), ``things I could move'' ($\to\!$\fieldname{movability\_prior}),
``where it stays still'' ($\to\!$\fieldname{motion\_class}$=$\textsc{static}). Highlighted instances
are the query result (tiny movers are ring-marked; dimmed $=$ not selected). The attributes are
object-level instances, not a per-voxel feature volume.}
\label{fig:repr}
\end{figure*}

\noindent\textbf{Notation.} The map is a set of object-instance elements
$\mathcal{E}=\{e_i\}$. Element $i$ is observed in keyframes through an instance mask
$\mathcal{M}_i\!\subset\!\Omega$ (pixel domain $\Omega$) and stores
\[
e_i=\big(\mathbf{p}_i,\ \mathbf{f}_i,\ o_i,\ \kappa_i,\ \rho_i,\ y_i,\ c_i\big),
\]
position $\mathbf{p}_i\!\in\!\mathbb{R}^3$, semantic feature $\mathbf{f}_i$, observed-motion score
$o_i\!\ge\!0$ with reliability $\kappa_i\!\in\![0,1]$, movability prior $\rho_i\!\in\![0,1]$, fused
class $y_i\!\in\!\mathcal{Y}\!=\!\{\textsc{static},\textsc{movable\_static},\textsc{moving}\}$, and
class confidence $c_i$. Camera intrinsics are $K$; the pose of keyframe $t$ is
$T_t=(R_t,\mathbf{t}_t)\in SE(3)$ (camera-to-world), known or estimated. Fig.~\ref{fig:repr}
shows the resulting map on real keyframes across three scenes: each object instance carries the
schema, and a natural-language query highlights the instances matching a single field.

\subsection{Semantic channel}
The semantic feature is an open-vocabulary \emph{region-CLIP} embedding: the full CLIP image
encoder~\cite{radford2021clip,ilharco2021openclip} is applied to the instance's masked crop,
\begin{equation}
\mathbf{f}_i=\mathrm{CLIP}_{\mathrm{img}}\!\big(\mathrm{crop}(I,\mathcal{M}_i)\big)\big/\big\|\!\cdot\!\big\|_2 ,
\end{equation}
which is markedly more discriminative than dense MaskCLIP~\cite{zhou2022maskclip} features and gives
a strong semantic-only baseline (the ConceptFusion/ConceptGraphs paradigm). A language
query $q$ scores element $i$ by cosine similarity
$s_i(q)=\mathbf{f}_i^{\!\top}\,\mathrm{CLIP}_{\mathrm{txt}}(q)$.

\subsection{Geometric observed-motion channel}
\begin{figure}[tbp]
\centering
\includegraphics[width=0.96\linewidth]{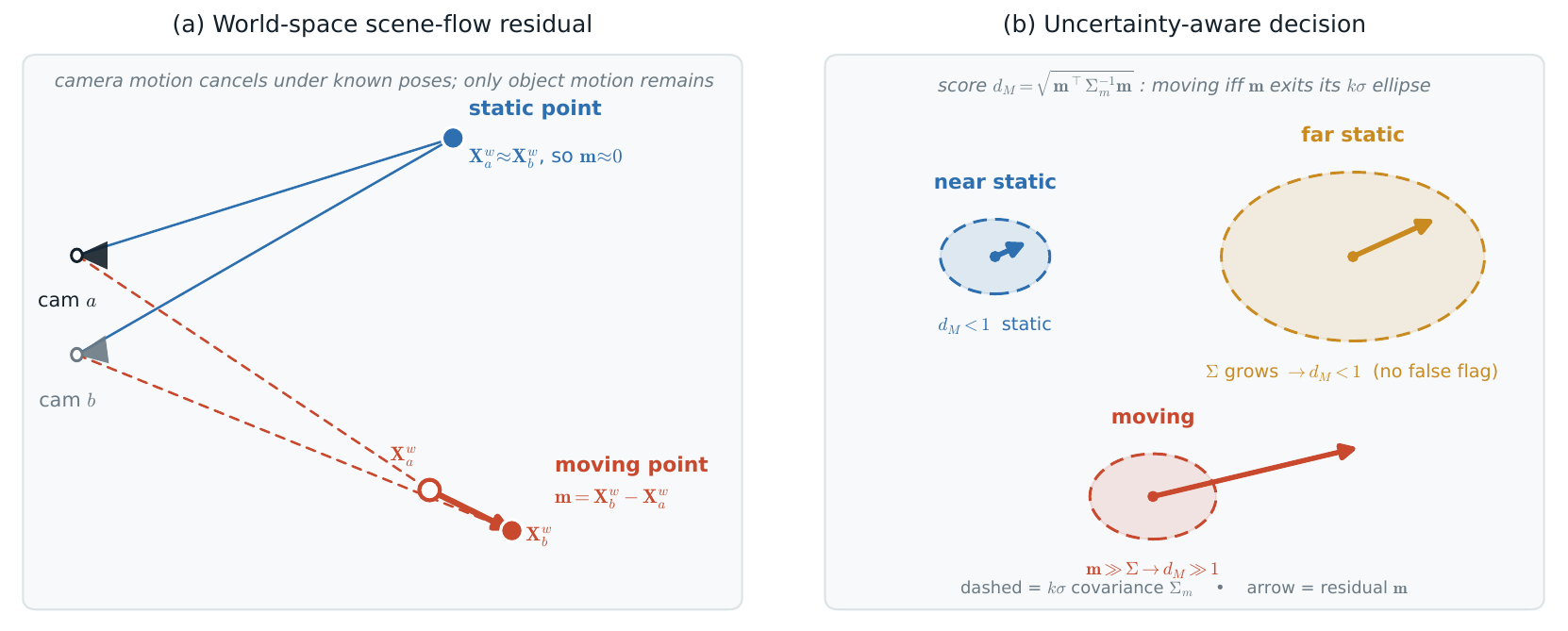}
\caption{Core mechanism. (a) With known/ego-refined poses, a \emph{static} surface point maps to
the same world coordinate from both views, so its residual $\mathbf{m}\!\approx\!0$; a \emph{moving}
point yields a residual equal to its world displacement. (b) Motion is scored by the Mahalanobis
magnitude of $\mathbf{m}$ against a range-dependent covariance $\Sigma_m$: far/noisy static points
get a larger $\Sigma_m$ and are \emph{not} false-flagged, while a true mover's residual exceeds its
$k\sigma$ ellipse.}
\label{fig:geometry}
\end{figure}
The principle (Fig.~\ref{fig:geometry}a): under known poses, only physical motion changes a surface
point's \emph{world} coordinate. For a pixel $u\!\in\!\Omega$ in keyframe $a$ with depth $z_a(u)$ we
back-project and lift to the world frame,
\begin{equation}
\mathbf{X}^{w}_a(u)=R_a\,\underbrace{z_a(u)\,K^{-1}[\,u_x,u_y,1\,]^{\!\top}}_{\text{camera-frame point}}+\mathbf{t}_a .
\end{equation}
Dense optical flow~\cite{teed2020raft} gives the correspondence $u'=u+\mathbf{w}_{a\to b}(u)$ in
keyframe $b$, with a forward--backward consistency mask $\mathcal{V}$ (we keep $u$ iff
$\lVert\mathbf{w}_{a\to b}(u)+\mathbf{w}_{b\to a}(u')\rVert<\epsilon$ and depth is valid at both
ends). The world-space \emph{scene-flow residual} is
\begin{equation}
\mathbf{m}(u)=\mathbf{X}^{w}_b(u')-\mathbf{X}^{w}_a(u).
\end{equation}

\noindent\textbf{Ego-refinement.} Pose error (e.g.\ from SLAM) adds a spurious rigid component to
every $\mathbf{m}$. We absorb it by re-estimating the residual relative transform from the dominant
(static) correspondences:
\begin{equation}
(\hat{R},\hat{\mathbf{t}})=\!\!\argmin_{R\in SO(3),\,\mathbf{t}}\sum_{u\in\mathcal{S}}\big\lVert R\,\mathbf{X}^w_a(u)+\mathbf{t}-\mathbf{X}^w_b(u')\big\rVert^2,
\end{equation}
where the inlier set $\mathcal{S}$ is found by RANSAC (moving points are outliers) and the fit is
closed-form: with inlier centroids $\bar{\mathbf{X}}_a,\bar{\mathbf{X}}_b$ and the SVD
$U\!S\,V^{\!\top}=\sum_{u\in\mathcal{S}}(\mathbf{X}^w_b(u')-\bar{\mathbf{X}}_b)(\mathbf{X}^w_a(u)-\bar{\mathbf{X}}_a)^{\!\top}$,
\begin{equation}
\hat{R}=U\,\mathrm{diag}\!\big(1,1,\det(UV^{\!\top})\big)\,V^{\!\top},\qquad
\hat{\mathbf{t}}=\bar{\mathbf{X}}_b-\hat{R}\,\bar{\mathbf{X}}_a,
\end{equation}
(Kabsch/Umeyama, the $\det$ term guarding against reflections). The refined residual is
$\tilde{\mathbf{m}}(u)=\mathbf{X}^w_b(u')-(\hat{R}\,\mathbf{X}^w_a(u)+\hat{\mathbf{t}})$.

\noindent\textbf{Uncertainty propagation.} The back-projection
$\mathbf{X}^{c}(u,z)=z\,K^{-1}[u_x,u_y,1]^{\!\top}$ is uncertain in both the flow correspondence and
the depth. Modelling these as independent zero-mean Gaussians with pixel variance
$\sigma_{\mathrm{px}}^2$ and a range-dependent axial depth variance
$\sigma_z^2(z)=\big(a+b\,(z-z_0)^2\big)^2$~\cite{nguyen2012noise}, a first-order (unscented-free)
propagation through the Jacobian $J=\partial\mathbf{X}^{c}/\partial(u_x,u_y,z)$ yields the
camera-frame covariance
\begin{equation}
\begin{aligned}
\Sigma^{c}&=J\,\mathrm{diag}\!\big(\sigma_{\mathrm{px}}^2,\sigma_{\mathrm{px}}^2,\sigma_z^2(z)\big)\,J^{\!\top},\\[1pt]
J&=\big[\,\tfrac{z}{f_x}\mathbf{e}_1,\ \tfrac{z}{f_y}\mathbf{e}_2,\ \tfrac{1}{z}\mathbf{X}^{c}\,\big],
\end{aligned}
\end{equation}
which, unlike a fixed diagonal model, correctly couples depth error into the \emph{lateral} world
axes for off-centre pixels. Rotating to the world frame ($\Sigma^{w}=R\,\Sigma^{c}R^{\!\top}$) and
treating the two endpoints as independent, the covariance of the refined residual
$\tilde{\mathbf{m}}=\mathbf{X}^w_b-(\hat R\,\mathbf{X}^w_a+\hat{\mathbf{t}})$ follows from
propagating $\Sigma^w_a$ through $\hat R$---the endpoint that the ego-refinement transforms---so
\begin{equation}
\Sigma_m=\Sigma^w_b+\hat R\,\Sigma^w_a\hat R^{\!\top}.
\label{eq:sigmam}
\end{equation}
Motion is scored by the Mahalanobis magnitude (Fig.~\ref{fig:geometry}b)
\begin{equation}
d_M(u)=\sqrt{\tilde{\mathbf{m}}(u)^{\!\top}\,\Sigma_m^{-1}\,\tilde{\mathbf{m}}(u)} .
\end{equation}
\textbf{Decision as a hypothesis test.} Under the null (static) hypothesis
$H_0\!:\mathbb{E}[\tilde{\mathbf{m}}]=\mathbf{0}$, the residual is zero-mean Gaussian, so
$d_M^2\sim\chi^2_3$ and the gate $d_M>\tau$ is a per-point likelihood-ratio test whose threshold
$\tau=\sqrt{\chi^2_{3,\,1-\alpha}}$ is fixed by a single static false-flag level $\alpha$
(e.g.\ $\alpha\!=\!0.05\Rightarrow\tau\!\approx\!2.80$). The tail probability
$p_{\mathrm{stat}}(u)=\Pr\!\big(\chi^2_3>d_M^2\big)$ is the static-consistency $p$-value (large when
the residual is explained by noise, small when it is not; we name it $p_{\mathrm{stat}}$ rather than
$p_{\mathrm{mot}}$ because it scores the \emph{static} hypothesis). Hence
far/noisy points inflate $\Sigma_m$, shrink $d_M$, and are \emph{not} false-flagged---the threshold
adapts to range instead of being a fixed metric distance.

\noindent\textbf{Per-object aggregation.} Over valid instance pixels
$\mathcal{P}_i=\mathcal{M}_i\cap\mathcal{V}$, the object score is the $75$th percentile of $d_M$ and
the reliability is the product of a fraction-moving term $\phi_i$ and a coherence term $\gamma_i$:
\begin{equation}
o_i=\mathrm{Q}_{75}\{d_M(u):u\in\mathcal{P}_i\},\qquad \kappa_i=\phi_i\,\gamma_i,
\label{eq:aggregation}
\end{equation}
\begin{align}
\phi_i &= \frac{1}{|\mathcal{P}_i|}\sum_{u\in\mathcal{P}_i}\mathbf{1}\!\left[d_M(u)>\tau\right], \\
\gamma_i &= \Bigg\lVert\, \frac{1}{|\mathcal{P}_i|}\sum_{u\in\mathcal{P}_i}\frac{\tilde{\mathbf m}(u)}{\lVert\tilde{\mathbf m}(u)\rVert} \,\Bigg\rVert ,
\end{align}
i.e.\ $\phi_i$ is the fraction of pixels above the motion-noise floor and $\gamma_i$ the directional
coherence (mean resultant length) of the residual vectors. Hence $\kappa_i$ is high only when a
coherent fraction of the instance moves together (a rigid mover), and low for boundary/flow bleed.

\noindent\textbf{Persistent-map update.} A sequence yields one observation per keyframe pair in which
an instance is visible with at least $30$ valid pixels (elements additionally require $\ge\!150$
pixels in some keyframe to be instantiated at all), so an element accumulates a set
$\mathcal{O}_i=\{(o_i^{(k)},\kappa_i^{(k)})\}_k$ rather than a single measurement. The stored
attributes are the elementwise medians over that set,
\begin{equation}
o_i \leftarrow \mathrm{median}_k\, o_i^{(k)},\qquad
\kappa_i \leftarrow \mathrm{median}_k\, \kappa_i^{(k)},
\label{eq:update}
\end{equation}
so a new observation is \emph{fused with} the history rather than overwriting it; the median is chosen
over a mean or a recency-weighted filter because single-pair outliers (flow failures, brief occlusion)
are the dominant error mode. The semantic feature is taken from the keyframe where the instance
subtends the most pixels. Consequently $o_i$ is an \emph{aggregate over the mapping episode}---``this
element has been observed to move''---and not an estimate of the instance's current velocity; a query
about the present state requires a fresh observation.

We are explicit about what this does \emph{not} include. Elements are keyed by instance
identity supplied by the segmentation front-end (ground-truth instance IDs in the AI2-THOR
experiments), so the association step itself is not a contribution of this work and is not evaluated
here; a deployed system would need geometric/appearance association, and errors there would propagate
into~\eqref{eq:update}. What we do claim is the fusion rule and the attribute it produces.

\subsection{Movability prior}
$\rho_i=\mathrm{VLM}(\ell_i)\in[0,1]$ answers ``can a \emph{$\ell_i$} move?'' for the object's
open-vocabulary label $\ell_i$, queried once per category and cached (no per-frame VLM cost).

\subsection{Fusion rule}
A confidence-aware rule maps the two channels to the class $y_i$; high-confidence \emph{observed}
motion overrides the prior, while unreliable/absent motion defers to it:
\begin{equation}
y_i=
\begin{cases}
\textsc{moving} & \text{if } o_i>\tau_o \,\wedge\, \kappa_i\ge\kappa_o,\\[2pt]
\textsc{movable\_static} & \text{else if } \rho_i\ge\rho_\tau,\\[2pt]
\textsc{static} & \text{otherwise}.
\end{cases}
\end{equation}
with class confidence $c_i=\kappa_i$ for \textsc{moving} and $c_i=\rho_i$ (resp.\ $1-\rho_i$) for
\textsc{movable\_static} (resp.\ \textsc{static}). The thresholds $(\tau_o,\kappa_o,\rho_\tau)$ make
the rule a small, swappable function.

\subsection{Query interface}
A natural-language query is parsed to a structured filter $\langle s,\ \mathit{field},\ \mathit{ret}\rangle$:
an optional semantic text $s$, a return type $\mathit{ret}$ (object or mask), and a single target
field that is one of \fieldname{observed\_motion}, \fieldname{movability\_prior},
\fieldname{motion\_class}, or none. The executor returns
\begin{equation}\label{eq:query}
\{\,e_i : s_i(s)\ge\theta_s \ \wedge\ \pi_{\mathit{field}}(e_i)\,\},
\end{equation}
where $\pi$ is the field predicate (e.g.\ $y_i\!=\!\textsc{moving}$ for ``the door that opens'',
$\rho_i\!\ge\!\rho_\tau$ for ``things I could move'', $y_i\!=\!\textsc{static}$ returning a mask for
``where it stays still''). Each intent thus touches exactly one field.

\noindent\textbf{The parser is a keyword intent router, and we specify it exactly.} The front-end is
not a language model and performs no compositional parsing; it is the following deterministic
procedure over the lower-cased query $t$. Three phrase sets are tested \emph{in this order}, and the
first match fixes the field (the order matters because \texttt{"could move"} contains
\texttt{"move"}):
{\small
\begin{itemize}\itemsep0pt \parskip0pt \leftskip-4pt
\item[(i)] $\mathcal{K}_{\text{prior}}$ (21 phrases: \texttt{could/can (i) move}, \texttt{movable},
\texttt{moveable}, \texttt{pick up}, \texttt{pickup}, \texttt{relocate}, \texttt{portable},
\texttt{can be moved}, \texttt{able to move}, \texttt{liftable}, \texttt{lift}, \texttt{carry},
\texttt{grab}, \dots) $\Rightarrow$ \fieldname{movability\_prior}, return objects;
\item[(ii)] $\mathcal{K}_{\text{static}}$ (19: \texttt{stays/stay still}, \texttt{stays put},
\texttt{static}, \texttt{fixed (in place)}, \texttt{immovable}, \texttt{does/doesn't move},
\texttt{not moving}, \texttt{in place}, \texttt{permanent}, \texttt{never moves},
\texttt{remains still}, \dots) $\Rightarrow$ \fieldname{motion\_class}$=$\textsc{static}, return mask;
\item[(iii)] $\mathcal{K}_{\text{moving}}$ (19: \texttt{opens}, \texttt{opening},
\texttt{that move(s)}, \texttt{is/are moving}, \texttt{in motion}, \texttt{actually
move/moving}, \texttt{just moved}, \texttt{currently moving}, \texttt{dynamic},
\texttt{walking}, \dots) $\Rightarrow$ \fieldname{observed\_motion}, return objects;
\item[(iv)] otherwise no motion filter: pure open-vocabulary semantic retrieval.
\end{itemize}}
\noindent The semantic text $s$ is the residue after deleting the matched phrases (longest-first) and
a fixed $62$-word stop list, plus the tokens \texttt{move/moves/moving}. The complete phrase and stop
lists are in the released code.

\noindent\textbf{Scope of the language claim.} This mechanism supports a \emph{closed set of three
intents over an open vocabulary of object nouns}: the noun phrase $s$ is matched by CLIP similarity
and is therefore open-vocabulary, while the motion intent is not---it is keyword routing. It has no
representation of negation, conjunction/disjunction, comparatives, quantifiers, or spatial relations,
and a query whose intent is phrased outside the three sets falls through to case~(iv) and silently
returns a semantic-only result. Accordingly we claim \emph{open-vocabulary object reference with
rule-based intent routing}, not natural-language understanding, and we mark NL querying \emph{partial}
for VLMM in Table~\ref{tab:capability}. An LLM front-end emitting the same
$\langle s,\mathit{field},\mathit{ret}\rangle$ filter is a drop-in replacement (an untested
implementation is included in the release); evaluating it is future work.

Because motion and movability are object-level properties, VLMM stores per-object elements rather
than a per-voxel feature volume; a query is therefore an attribute filter over the element set
(Eq.~\ref{eq:query}), not a nearest-neighbour lookup over a feature grid as in voxel-based
open-vocabulary maps. Fig.~\ref{fig:repr} visualizes the resulting per-object attributes projected
onto real keyframes; for grid-based planners they can additionally be rasterized into an occupancy
grid on demand.

\section{Results \& Discussion}\label{sec:results}

\subsection{Implementation and compute}
The full system is implemented in \textbf{Python}~3.10 with PyTorch~2.5. The semantic channel uses
open\_clip ViT-B/16~\cite{ilharco2021openclip}; the observed-motion channel uses RAFT optical
flow~\cite{teed2020raft}; a YOLOv8-seg model supplies instance masks on real RGB-D; and the embodied
study runs in the AI2-THOR~\cite{kolve2017ai2thor} simulator. All experiments run on a
\textbf{single laptop-class GPU (NVIDIA RTX~4060, 8\,GB)}---an explicit design constraint. To stay
within 8\,GB we (i) cache per-keyframe dense CLIP features to disk so peak VRAM never holds all
frames at once, (ii) treat the VLM/LLM movability prior as a cached per-category call (no per-frame
cost), and (iii) avoid any component that needs a multi-GPU server.

\subsection{Datasets}
\textbf{Simulator (exact ground truth).} AI2-THOR provides, per object, an instance mask, movability
flags (\texttt{pickupable}/\texttt{moveable}), and exact motion (objects are displaced by known
deltas). We render three genuinely different scene types---kitchen, living room, bedroom (67
observed objects)---and use them for the field-necessity ablation because every label is exact;
camera geometry was validated to $1.1$\,cm cross-view static-point alignment.
\textbf{Real RGB-D.} TUM RGB-D~\cite{sturm2012tum} (\texttt{walking\_static}, \texttt{walking\_xyz},
\texttt{walking\_rpy}, \texttt{desk\_with\_person}) and the Bonn dynamic
dataset~\cite{palazzolo2019bonn} (\texttt{crowd}, \texttt{person\_tracking}) give six real noisy
dynamic sequences across two labs/sensors; moving-region ground truth comes from a person segmenter
(a proxy; see Limitations). Note the AI2-THOR scenes contain \emph{no people}---every mover is a
household object (Kettle, Pan, Book, Laptop, Bowl, Tissue~Box)---so the motivating non-person
``door that opens'' query is evaluated directly with exact, non-proxy ground truth (Q\_moving,
Table~\ref{tab:ablation}); the person proxy is only the real-data stand-in.

\subsection{Evaluation metrics and parameters}
We report \emph{average precision} (AP) and precision/recall/F1 for moving-vs-static detection and
for the attribute queries---not accuracy, since true negatives dominate---mask IoU for region
queries, and \emph{expected calibration error} (ECE)~\cite{guo2017calibration} for confidence. Key
settings, fixed across all scenes: region-CLIP crops at $640$\,px; RAFT with a $2$\,px
forward--backward consistency gate; the depth-noise model~\cite{nguyen2012noise}
$(a,b,z_0)\!=\!(1.2,1.9,0.4)\!\times\!10^{-3}$ with lateral $\sigma_{\mathrm{px}}\!=\!1.5$;
ego-refinement by RANSAC ($2$\,cm inlier threshold) and Kabsch, enabled on the real sequences
(estimated poses) and disabled in the exact-GT simulator, where poses are exact and refinement would
absorb genuine object motion---there $\hat R\!=\!I$ and~\eqref{eq:sigmam} reduces to
$\Sigma^w_a\!+\!\Sigma^w_b$; per-object score
$o_i=\mathrm{Q}_{75}(d_M)$; frame gap $k$ chosen per dataset; and a single fixed fusion threshold
triple $(\tau_o,\kappa_o,\rho_\tau)$. Every reported number traces to one pinned run per experiment.

\noindent\textbf{Choice of the aggregation quantile.} $\mathrm{Q}_{75}$ in~\eqref{eq:aggregation} is a
robustness compromise: a low quantile is dominated by the stationary majority of a partially-moving
object (an opening door is mostly hinge), while the maximum is set by the single noisiest pixel.
Sweeping $p$ with the rest of the pipeline fixed, the moving-vs-static ROC-AUC on the exact-GT maps is
$1.000$ for every $p\!\in\![50,90]$, $0.995$ at $p\!=\!95$, and falls to $0.923$ (best-F1 $0.571$) at
$p\!=\!100$. The decision is therefore insensitive to $p$ over a broad interior band and only the
maximum is clearly harmful; $75$ is a representative interior choice rather than a tuned one.
(The sweep re-runs the motion pipeline rather than reusing the stored maps; at $p\!=\!75$ it
reproduces the perfect mover/non-mover separation reported under \emph{Threshold sensitivity} below,
though the fastest mover's score shifts by ${\sim}5\%$ under a different optical-flow build---the
margin is wide enough that the conclusion is unaffected.)
We note the limits of this sweep honestly: with $6$ movers among $67$ objects the exact-GT study is
saturated, so the result demonstrates \emph{insensitivity}, not optimality. It also cannot resolve the
interaction with apparent object size that a quantile might be expected to have---all six movers fall
in the two smaller size terciles, leaving the large-object AUC undefined. Establishing that
interaction needs the larger and more varied dynamic evaluation discussed in Sec.~\ref{sec:limits}.

\noindent\textbf{Threshold sensitivity.} The reported AP and false-flag numbers are
\emph{rank-based} and therefore independent of the fusion thresholds; only the discrete
\fieldname{motion\_class} assignment depends on $(\tau_o,\kappa_o,\rho_\tau)$. On the exact-GT maps
(67 objects) observed motion cleanly separates movers ($d_M\!\in\![0.69,3.65]$) from non-movers
($d_M\!\le\!0.05$), so the motion gate is insensitive across an order of magnitude
$\tau_o\!\in\![0.05,0.69]$ (3-class macro-F1 constant at $0.905$), degrading only once $\tau_o$
exceeds the slowest mover's score. The movability gate is more sensitive: macro-F1
$\in[0.905,0.966]$ over $\rho_\tau\!\in\![0.40,0.60]$ and falls to $0.77$ by $\rho_\tau\!=\!0.70$;
our fixed $\rho_\tau\!=\!0.55$ is conservative, not tuned (lower values score \emph{higher}). We thus
place $\tau_o$ in the wide observed-motion gap and report the $\rho_\tau$ dependence openly.

\subsection{Schema fields are non-substitutable (exact-GT ablation)}
\begin{figure*}[t]
\centering
\includegraphics[width=0.86\textwidth]{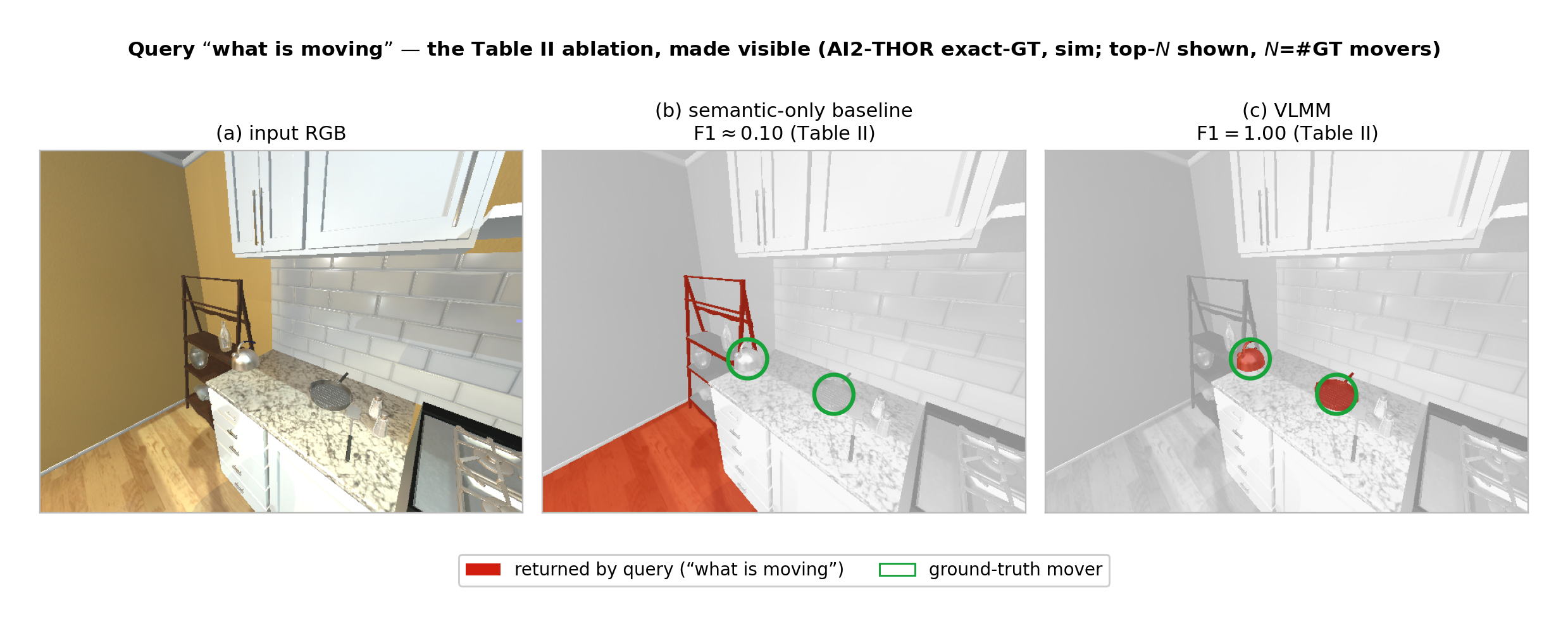}
\vspace{-10pt}
\caption{\footnotesize Query ``what is moving'' made visible (one THOR scene, exact-GT sim;
top-$N$ shown, $N\!=\!\#$GT movers). The semantic-only baseline ranks by CLIP similarity to ``an
object that is currently moving'' (Table~\ref{tab:ablation} row~i, $\mathrm{F1}\!\approx\!0.10$) and
returns the wrong instances; VLMM ranks by \fieldname{observed\_motion} and returns exactly the
ground-truth movers (green rings). Built from the same pinned map as Table~\ref{tab:ablation}.}
\label{fig:contrast}
\end{figure*}

Table~\ref{tab:ablation} reports AP on three THOR scenes. We read it as a \emph{non-substitutability}
check, not a discovery of necessity: since a query routes to a single field, the diagonal (a query
losing its own field) is partly definitional. The informative content is therefore (a) whether
\emph{strong semantics alone} can answer the motion queries, and (b) the \emph{off-diagonal}---whether
one motion field can substitute for the other. Both are negative. Semantic-only fails the motion
query ($0.10$, near the $0.09$ random rate) \emph{even with strong region-CLIP features}: image
semantics do not encode observed motion, so the failure is fundamental, not a weak-feature artifact.
The two motion channels are not interchangeable: the prior alone scores only $0.30$ on ``what is
moving'' (movability $\neq$ observed motion), while observed-motion alone collapses ``what could
move'' and ``stays still'' (an unmoved movable object reveals no motion). Only the full schema
answers all three. The uncertainty weighting ties here on exact depth (rows~iv and~v) and separates
only under sensor noise (Table~\ref{tab:unc})---it is a real-noise property, not a clean-sim one.
Fig.~\ref{fig:contrast} makes this concrete: on ``what is moving'' the semantic-only baseline
returns the wrong instances while VLMM, ranking by \fieldname{observed\_motion}, returns exactly the
ground-truth movers.

\noindent\textbf{Measured baselines.} The two dominant prior designs, reimplemented on the
\emph{same} exact-GT data, both fail the motion query. A \emph{VLMaps/ConceptFusion-style} CLIP map
(per-element features by text similarity), queried with the best of six phrasings (incl.\ ``the door
that opens''), reaches only AP $=0.18$ on ``what is moving'' (random $0.09$) versus VLMM's $1.00$---
appearance features do not encode instantaneous motion---though it partially answers ``what could
move'' ($0.66$), since movability is semantic. \emph{DualMap}'s anchor-vs-volatile mechanism (CLIP
similarity to a static-category template) answers ``what could move'' well ($0.80$) but
\emph{conflates moving with movable}: only $0.32$ on ``what is moving,'' as an anchor/volatile state
cannot isolate the currently-moving subset. Both lack the observed-motion channel VLMM adds; these
are the documented \emph{mechanisms}, not the released systems (Limitations).

\begin{table}[t]
\centering
\caption{\footnotesize Schema-field non-substitutability ablation (AP, exact-GT sim, three THOR scenes).}
\label{tab:ablation}
\footnotesize
\setlength{\tabcolsep}{4pt}
\resizebox{0.98\columnwidth}{!}{
\begin{tabular}{lccc}
\toprule
map configuration & Q\_moving & Q\_movable & Q\_static\\
\midrule
(i) semantic-only [region-CLIP] & 0.10 & 0.40 & 0.70\\
(i) semantic-only [MaskCLIP]    & 0.14 & 0.38 & 0.44\\
(ii) + observed motion only     & \textbf{1.00} & 0.73 & 0.71\\
(iii) + movability prior only   & 0.30 & \textbf{0.99} & \textbf{0.99}\\
(iv) full fusion + uncertainty  & \textbf{1.00} & \textbf{0.99} & \textbf{0.99}\\
(v) full fusion, no uncertainty & 1.00 & 0.99 & 0.99\\
random / prevalence             & 0.09 & 0.48 & 0.52\\
\bottomrule
\end{tabular}}
\end{table}

\subsection{Uncertainty is the differentiator (multi-scene)}
The uncertainty channel separates VLMM from Dewan'17. Table~\ref{tab:unc} shows that switching the
raw world-displacement score to the Mahalanobis score improves moving-vs-static AP in all six real
sequences (mean $+0.10$) and reduces the far-static false-flag rate in all six. On exact-GT sim,
the two are tied on noise-free depth (nothing to be uncertain about) but diverge by $+0.27$ AP once
we \emph{inject} a Kinect-quadratic depth noise $\sigma=0.0012+0.0028\,z^2$ (Table~\ref{tab:unc}, last rows).
We stress that this injected noise is \emph{deliberately different} from the method's assumed
covariance (the Nguyen axial model of Sec.~III-C): the method only \emph{approximates} the true
sensor noise, so the test is fair rather than self-confirming: uncertainty is neutral on clean depth
and essential under real sensor noise.

\begin{table}[t]
\centering
\caption{\footnotesize Uncertainty necessity: raw vs.\ Mahalanobis (AP, far-static false-flag rate).}
\label{tab:unc}
\footnotesize
\setlength{\tabcolsep}{4pt}
\resizebox{0.98\columnwidth}{!}{
\begin{tabular}{lccc}
\toprule
scene & AP raw & AP maha & far-static FF (raw$\rightarrow$maha)\\
\midrule
TUM walking\_static & 0.19 & 0.32 & 98.5\%\,$\rightarrow$\,13.4\%\\
TUM walking\_xyz    & 0.20 & 0.33 & 99.6\%\,$\rightarrow$\,44.3\%\\
TUM walking\_rpy    & 0.14 & 0.24 & 67.2\%\,$\rightarrow$\,64.9\%\\
TUM desk\_w\_person & 0.10 & 0.13 & 81.6\%\,$\rightarrow$\,50.0\%\\
Bonn crowd          & 0.30 & \textbf{0.48} & 39.5\%\,$\rightarrow$\,\textbf{8.2\%}\\
Bonn person\_track  & 0.08 & 0.13 & 62.8\%\,$\rightarrow$\,43.1\%\\
\midrule
THOR exact-GT, clean & 1.00 & 1.00 & (tied --- no noise)\\
THOR exact-GT, Kinect noise & 0.41 & \textbf{0.68} & $+0.27$ gap\\
\bottomrule
\end{tabular}}
\end{table}

\subsection{Calibration}
The raw per-element motion confidence is \emph{not} calibrated: ECE $=0.30$ (pooled, 4.8M pixels),
overconfident. It is rank-useful (it drives the $+0.10$ AP). A standard post-hoc isotonic
calibration, fit on TUM and tested held-out on Bonn, reaches ECE $=0.10$ (Fig.~\ref{fig:calpose}a,b).
We therefore claim ``uncertainty-weighted, post-hoc calibratable,'' \emph{not} ``well-calibrated.''

\subsection{Robustness to estimated poses}
\begin{figure*}[t]
\centering
\includegraphics[width=0.86\textwidth]{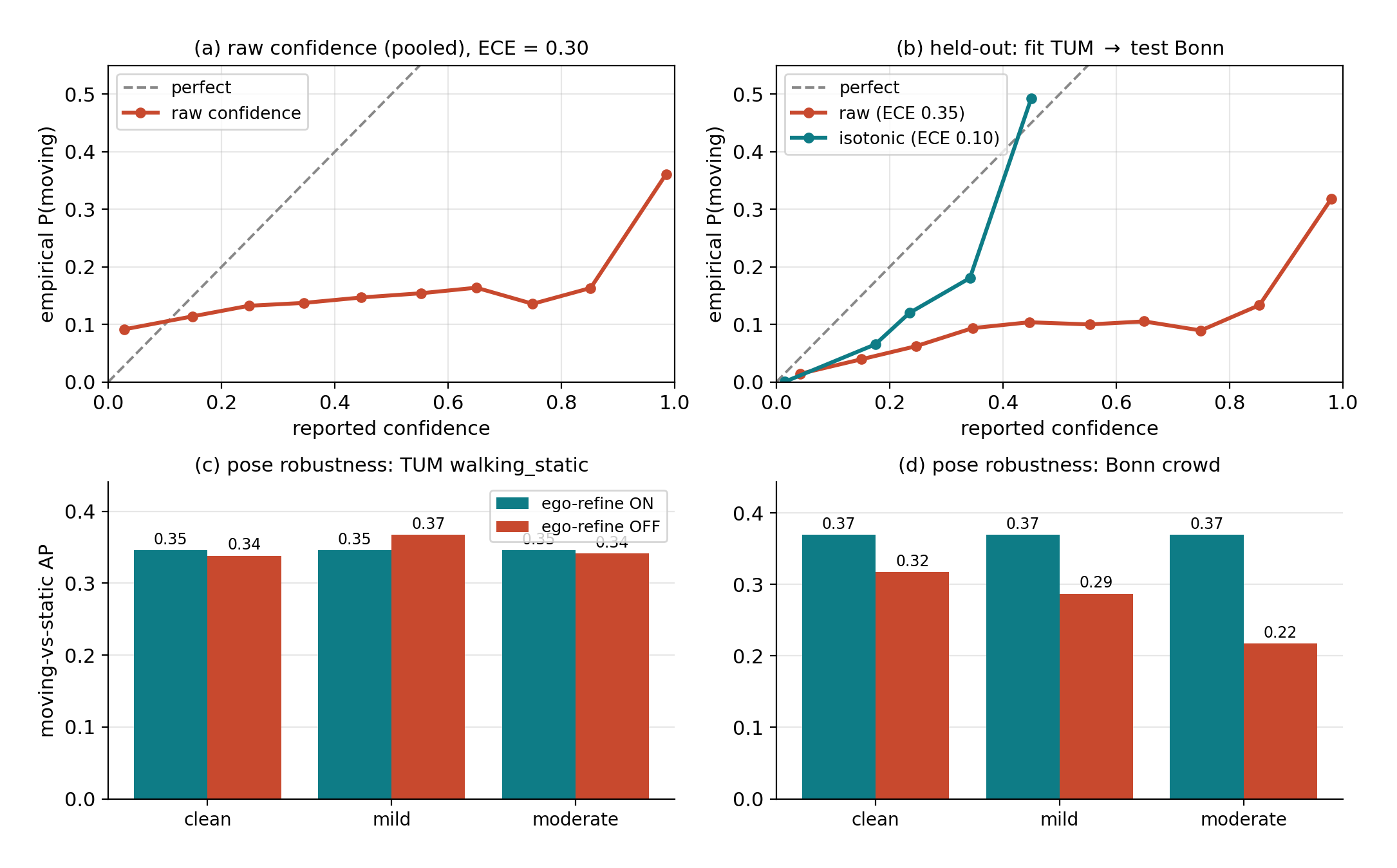}
\vspace{-10pt}
\caption{\footnotesize Calibration and pose robustness. (a)~raw motion confidence is uncalibrated
(ECE $0.30$); (b)~post-hoc isotonic calibration, fit on TUM and tested held-out on Bonn, reaches
ECE $0.10$. (c,d)~ego-refinement makes moving-vs-static AP invariant to estimated-pose error;
without it AP degrades.}
\label{fig:calpose}
\end{figure*}

The observed-motion channel assumes known poses. Fig.~\ref{fig:calpose}c,d injects per-frame pose
error: with ego-refinement the AP is \emph{invariant} to pose error (it re-derives the relative pose
from static structure), while without it AP degrades (Bonn $0.32\!\rightarrow\!0.22$ at
$1.5^\circ/3$\,cm). The channel therefore does not require accurate SLAM poses (assuming a
dominantly-static scene).

\subsection{Query interface and parser}
The three canonical queries plus paraphrases route to the correct field; a confidence threshold on
``reliably static'' trades recall for precision ($0.81\!\rightarrow\!1.00$). On 30 held-out
paraphrases the rule parser routes $24/30=80\%$ correctly, failing on negation---confirming it is
intent routing, not language understanding (an LLM parser is the camera-ready fix).

\vspace{-10pt}
\section{Limitations}\label{sec:limits}
(1) The strong ablation numbers are exact-GT \emph{simulation}; real-world object-level F1 is
$\approx0.5$ (motion-bleed; flow under-measures small/fast objects)---we claim relative, not SOTA,
results. (2) Confidence is uncertainty-weighted, not raw-calibrated. (3) The query parser is intent
routing, not NLU. (4) Real-data motion GT is a person segmenter, conflating ``moving'' with
``person'' (static people are false positives, non-person movers missed); the real-data numbers
(Table~\ref{tab:unc}) thus establish the uncertainty channel's effect under sensor noise, \emph{not}
general non-person detection---which we evaluate only in simulation (exact GT, incl.\ injected
Kinect-level noise, $+0.27$ gap). A real non-person-mover benchmark is the main remaining gap.
(5) We measure VLMaps- and DualMap-\emph{style} baselines (documented mechanisms, shared data;
Sec.~\ref{sec:results}); the \emph{full released} systems remain. (6) Motion thresholds need
per-sensor calibration; the ranking is the scale-robust quantity.
\vspace{-10pt}
\section{Conclusion}
VLMM adds a queryable, uncertainty-aware, fused motion attribute to open-vocabulary 3D maps. The
combination is novel against the closest prior art; the schema fields are shown non-substitutable on
exact-GT multi-scene simulation; and the uncertainty channel---the differentiator from prior
fused-motion work---is validated across six real noisy sequences and under realistic sensor noise in
simulation. Calibration and the scope of the claims are reported in full, leaving a real
non-person-mover benchmark and a released-baseline comparison as future work.

\bibliographystyle{IEEEtran}
\bibliography{refs}
\end{document}